\documentclass[letterpaper, 10 pt, conference]{ieeeconf}  %

\IEEEoverridecommandlockouts                              %

\usepackage{graphicx} 

\usepackage{amsmath,amssymb,amsfonts}
\usepackage{cite}
\usepackage{multirow}
\usepackage{algorithmic}
\usepackage[ruled,vlined,lined,linesnumbered,boxed,commentsnumbered]{algorithm2e}
\usepackage{booktabs}
\usepackage[caption=false]{subfig}
\usepackage{color}
\usepackage{balance}
\usepackage{threeparttable}
\usepackage{rotating}
\usepackage{pifont}
\usepackage[dvipsnames]{xcolor}
 
\definecolor{f5}{gray}{1}
\definecolor{f4}{gray}{0.8}
\definecolor{f3}{gray}{0.6}
\definecolor{f2}{gray}{0.4}
\definecolor{f1}{gray}{0.2}
\definecolor{f0}{gray}{0}

\title{\LARGE \bf
Characterizing SLAM Benchmarks and Methods for the Robust Perception Age
}

\author{Wenkai Ye$^{1}$, Yipu Zhao$^{1}$, and Patricio A. Vela$^{1}$%
\thanks{$^{1}$%
Wenkai Ye, Yipu Zhao, and Patricio A. Vela  
are with School of Electrical and Computer Engineering, 
Georgia Institute of Technology, Atlanta, Georgia, USA. 
{\tt\small \{wye35,yzhao347,pvela\}@gatech.edu}. }%
\thanks{
This work was supported in part by the China Scholarship Council 
(CSC Student No: 201606260089) and the National Science Foundation 
(Award \#1816138).
}%
}

\usepackage{url}
\usepackage{slam_macros}

\graphicspath{ {./images/}}

\begin{document}

\maketitle
\thispagestyle{empty}
\pagestyle{empty}

\begin{abstract}
The diversity of SLAM benchmarks affords extensive testing of SLAM
algorithms to understand their performance, individually or in relative
terms. The \textit{ad-hoc} creation of these benchmarks does not necessarily
illuminate the particular weak points of a SLAM algorithm when performance is
evaluated. In this paper, we propose to use a decision tree to identify challenging 
benchmark properties for state-of-the-art SLAM algorithms and important components 
within the SLAM pipeline regarding their ability to handle these challenges.
Establishing what factors of a particular
sequence lead to track failure or degradation relative to these
characteristics is important if we are to arrive at a strong 
understanding for the core computational needs of a robust SLAM algorithm.
Likewise, we argue that it is important to profile the computational
performance of the individual SLAM components for use when benchmarking.
In particular, we advocate the use of time-dilation during ROS bag
playback, or what we refer to as \textit{slo-mo} playback. 
Using \textit{slo-mo} to benchmark SLAM instantiations can provide clues to how
SLAM implementations should be improved at the computational component
level. Three prevalent VO/SLAM algorithms and two low-latency algorithms 
of our own are tested on selected typical sequences, which are generated from 
benchmark characterization, to further demonstrate the benefits achieved from 
computationally efficient components.
\end{abstract}

\section{Introduction}

Simultaneous localization and mapping (SLAM) is a core computational
component supporting several application scenarios in augmented/virtual
reality and autonomous robotics. As such, benchmarks for assessing the
performance of SLAM reflect the diverse deployment profiles of potential
application scenarios.  
They reflect a wide variety of 
sensors \cite{smith2009new,KITTI,RGBDbenchmark}, 
platforms \cite{zhao2019closednav,delmerico2018benchmark},
motion patterns \cite{urban2017lafida,schubert2018vidataset,burri2016euroc}, 
scene properties \cite{ICLNUIM,maddern20171}, 
and other meaningful characteristics. 
Given that a large portion of benchmarks are empirical, specialized (to
use case) scenarios recorded for evaluation through replay, there can be
a lack of control over important configuration variables related to
performance.
Furthermore, the diversity of software interfaces for the different datasets
and algorithms complicates comprehensive evaluation.
SLAMBench \cite{nardi2015introducing} addresses this last issue through the use of a
common API for evaluating algorithms with an emphasis on analysis of
different computational platforms and run-time parameters.
SLAMBench performance metrics include energy consumption, accuracy, and
computational performance. 
Follow-up work, SLAMBench 2~\cite{bodin2018slambench2}, improves API consistency 
and includes several SLAM implementations modified for compatibility with the
API.  The input stream can be synthetic data generated from simulations
\cite{ICLNUIM,li2018interiornet}. Simulation addresses the earlier point through the creation
of controlled scenarios that can be systematically perturbed or modified.
We hope to advance the practice of benchmarking by providing a meta-analysis
or design of experiments inspired analysis of the SLAM benchmarks and
algorithms with respect to accuracy and computation.  The analysis will
characterize existing benchmarks and identify critical components of the SLAM
pipeline under different benchmark characteristics. 

Our contributions in this direction follows in the subsequent sections.
Section \ref{sec:bench} lists existing benchmarks and briefly describes their
characteristics according to properties known to impact SLAM accuracy.
Analysis of differentiating factors regarding difficulty level is performed
to arrive at dominant factors influencing the difficulty annotation.
Section \ref{sec:timing} reviews time profiling outcomes of SLAM
instantiations in order to determine the time allocation required for
(sufficiently) complete execution of the SLAM pipeline prior to receipt of the 
next frame.  Providing sufficient time for the computations enables
separating the latency factor from the algorithm factor for establishing the
limiting bound of accuracy performance fo SLAM instances relative to existing
benchmarks.
Section \ref{sec:eval} applies 
three prevalent visual SLAM
algorithms and two low-latency counterparts
to a 
balanced
benchmark set as determined by the analysis of Section \ref{sec:bench}. The aim
of the study is to confirm that the qualitative assessment matches the
quantitative outcomes with the latter annotations determined by the
accuracy results.
The outcome distribution will be clustered into four performance classes:
\textit{fail}, \textit{low}, \textit{medium}, and \textit{high}, based on 
clustering the accuracy outcomes into three equal density regions plus
adding a fail category.  Comparison of the resulting decision trees will
establish whether the primary factors impacting performance relative to the
distinct performance categories are consistent or if a different
prioritization is in order.
The described analysis should provide a means to establish where structural
weaknesses of published SLAM methods lie and where future research effort
should be dedicated to maximize impact. %
The emphasis will be on monocular SLAM as improvements to monocular systems
should translate to the same for stereo and visual-inertial implementation
\cite{zhao2019closednav,zhao2019tro}.
We anticipate that the findings will support more systematic study of SLAM
algorithms in this new era of SLAM research, dubbed the ``Robust
Perception Age'' \cite{cadena2016past}.

\begin{table*}[t]
\centering
\caption{Characterization of Selected Sequence Properties}
\begin{tabular}{l|c|c|c|c|c|c|c}
\toprule
\midrule
\textbf{Sequence} & \textbf{Platform} & \textbf{Scene ($x_1$)} & \textbf{Duration ($x_2$)} & \textbf{Motion Dyn. ($x_3$)} & \textbf{Environ. Dyn. ($x_4$)} & \textbf{Revisit Freq. ($x_5$)} & \textbf{Difficulty ($y$)} \\
\midrule[0.1pt]
\textit{Seq 04}~\cite{KITTI} & Car & Outdoor & Short & Low & High & Low & Easy \\
\midrule[0.1pt]
\textit{lr kt0}~\cite{ICLNUIM} & Synthesized & Indoor & Short & Low & Low & Low & Easy \\
\midrule[0.1pt]
\textit{f2 desk person}~\cite{sturm12iros_ws} & HandHeld & Indoor & Short & Low & Medium & Low & Easy \\
\midrule[0.1pt]

\textit{Conf. hall2} & AR Headset & Indoor & Medium & Low & Medium & High & Medium \\
\midrule[0.1pt]
\textit{Seq 02}~\cite{KITTI} & Car & Outdoor & Medium & High & Medium & Low & Medium \\
\midrule[0.1pt]
\textit{room3}~\cite{schubert2018vidataset} & HandHeld & Indoor & Short & High & Low & Low & Medium \\
\midrule[0.1pt]
\textit{of kt3}~\cite{ICLNUIM} & Synthesized & Indoor & Short & Low & Low & Low & Medium \\
\midrule[0.1pt]

\textit{MH 05 diff}~\cite{burri2016euroc} & MAV & Indoor & Short & Medium & Low & High & Difficult \\
\midrule[0.1pt]
\textit{V1 03 diff}~\cite{burri2016euroc} & MAV & Indoor & Short & High & Low & High & Difficult \\
\midrule[0.1pt]
\textit{Corridor} & AR Headset & Indoor & Medium & Medium & Medium & High & Difficult \\
\midrule[0.1pt]
\textit{NewCollege}~\cite{smith2009new} & Round Robot & Outdoor & Long & Medium & Medium & High & Difficult \\
\midrule[0.1pt]
\textit{outdoor4}~\cite{schubert2018vidataset} & HandHeld & Outdoor & Long & Medium & Medium & Low & Difficult \\
\midrule[0.1pt]
\bottomrule
\end{tabular} 
\label{tab:summ_vslam}
\end{table*}

\section{Benchmark Properties}
\label{sec:bench}

Benchmarking for SLAM varies based on evaluation choices made by different
research teams. Some prioritize a select set of benchmark datasets based on
anticipated deployment characteristics \cite{delmerico2018benchmark}. 
Others seek to understand and confirm the general performance properties of
a set of methods \cite{li2016experimental}, or to explore the solution
landscape associated to parametric variations of a single strategy
\cite{nardi2015introducing}. Our interest is in understanding the general
performance landscape and what subset of available datasets could be used
to evaluate general deployment scenarios. If such a subset were to exist,
computed averages of the quantitative outcomes could provide a common
metric with which to score and compare the impact of algorithmic choices in
SLAM implementations.

We performed a literature search for benchmark datasets associated to SLAM
algorithms, and any other visual sequence data with ground truth pose
information permitting quantitative evaluation of camera pose versus time.
Published benchmarks for which the data is no longer
available were excluded, such as Rawseeds\cite{fontana2014rawseeds}.
In the end, the following corpus of benchmark datasets was identified:  
NewCollege\cite{smith2009new},				%
Alderley\cite{milford2012seqslam},			%
Karlsruhe\cite{Geiger2010ACCV}, 			%
Ford Campus\cite{pandey2011ford},			%
Malaga 2009\cite{malaga2009},				%
CMU-VL\cite{badino2011visual}, 			%
TUM RGBD\cite{sturm12iros_ws}, 				%
KITTI\cite{KITTI}, 							%
Malaga Urban\cite{blanco2014malaga},		%
ICL NUIM\cite{ICLNUIM}, 					%
UMich NCLT\cite{carlevaris2016university},	%
EuRoC\cite{burri2016euroc},				%
Nordland\cite{sunderhauf2013we},			%
TUM Mono\cite{engel2016monodataset},		%
PennCOSYVIO\cite{pfrommer2017penncosyvio}, %
Zurich Urban MAV\cite{majdik2017zurich}, 	%
RobotCar\cite{maddern20171},				%
TUM VI\cite{schubert2018vidataset},		%
BlackBird\cite{antonini2018blackbird}, 	%
and a Hololens benchmark of our own. 		%
Altogether, they
reflect over 310 sequences with available ground truth signals.

\begin{figure*}[t]
  \centering
\subfloat[KITTI \textit{Seq. 02}\label{fig:kitti02}]
        {\includegraphics[height=1in]{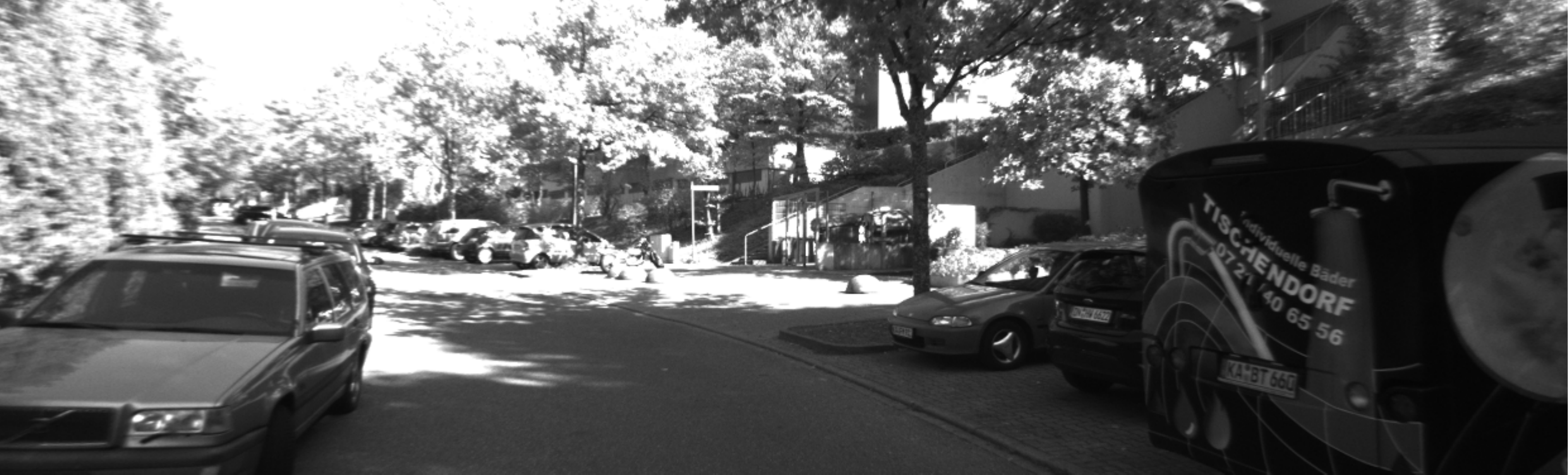}}
    \quad
\subfloat[TUM VI \textit{room3}\label{fig:room3}]
        {\includegraphics[height=1in]{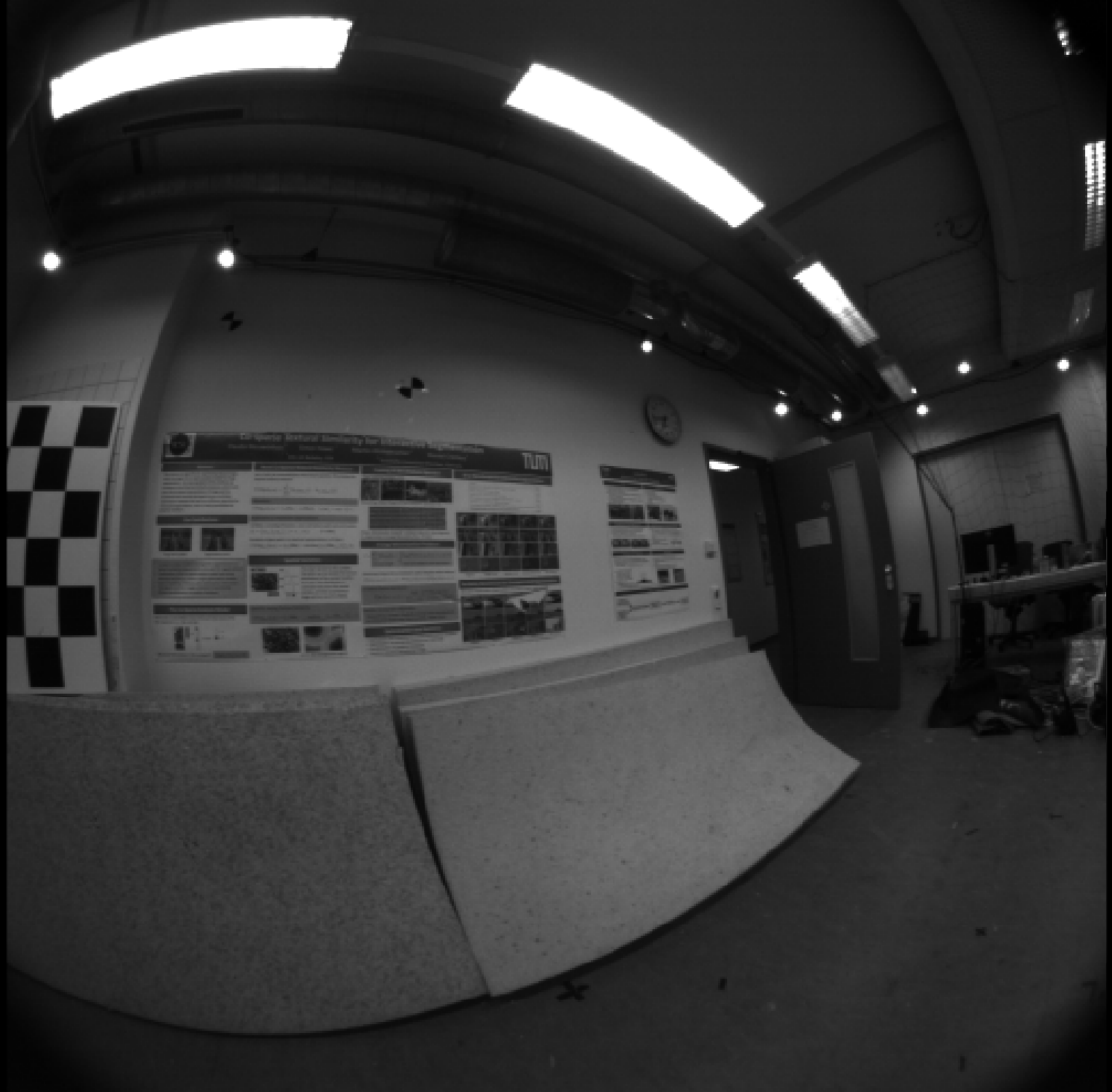}}
    \quad
\subfloat[TUM VI \textit{outdoor4}\label{fig:outdoor4}]
        {\includegraphics[height=1in]{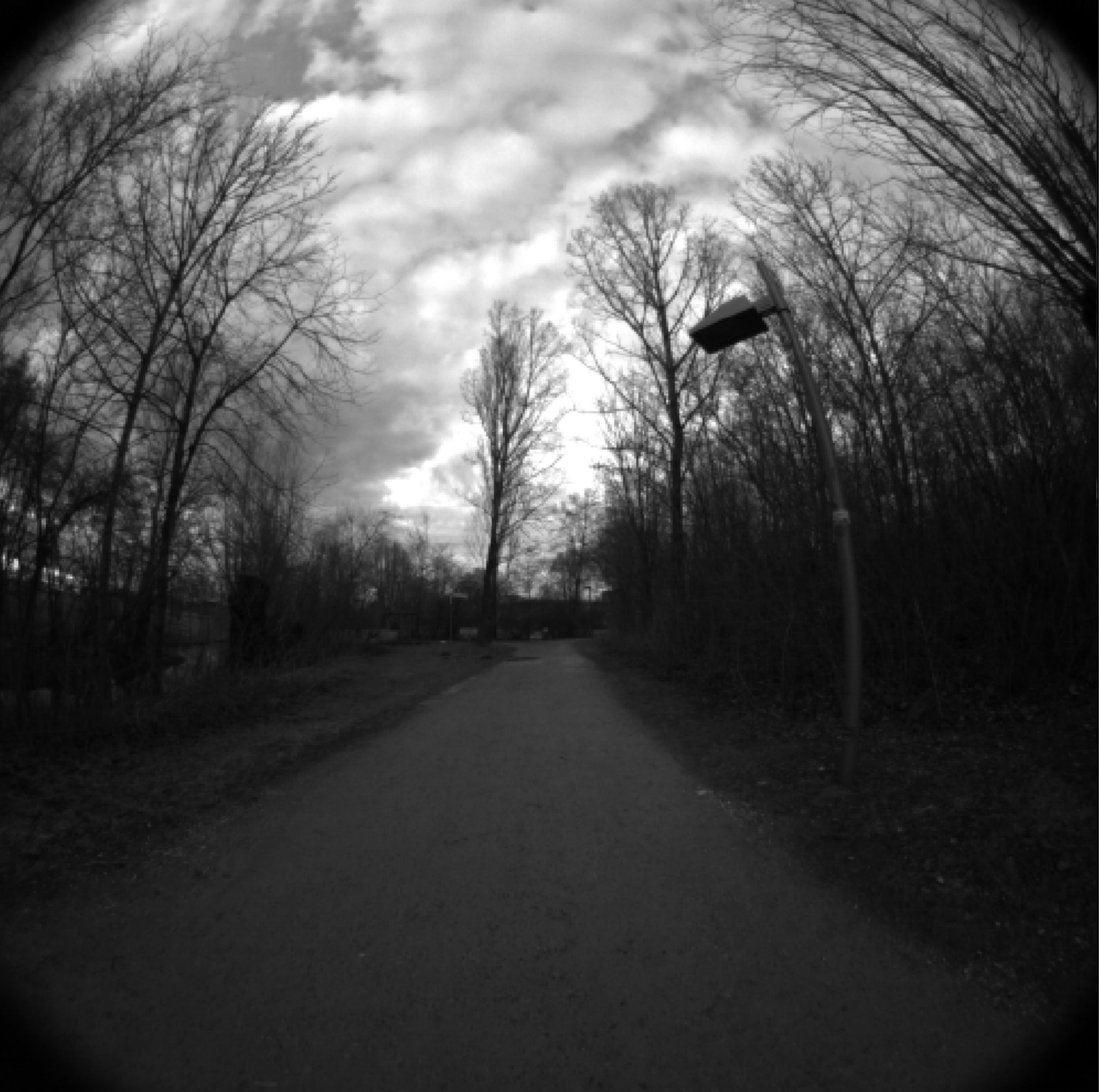}}
            \quad
\subfloat[\textit{Conf. Hall1} \label{fig:mean and std of net24}]
        {\includegraphics[height=1in]{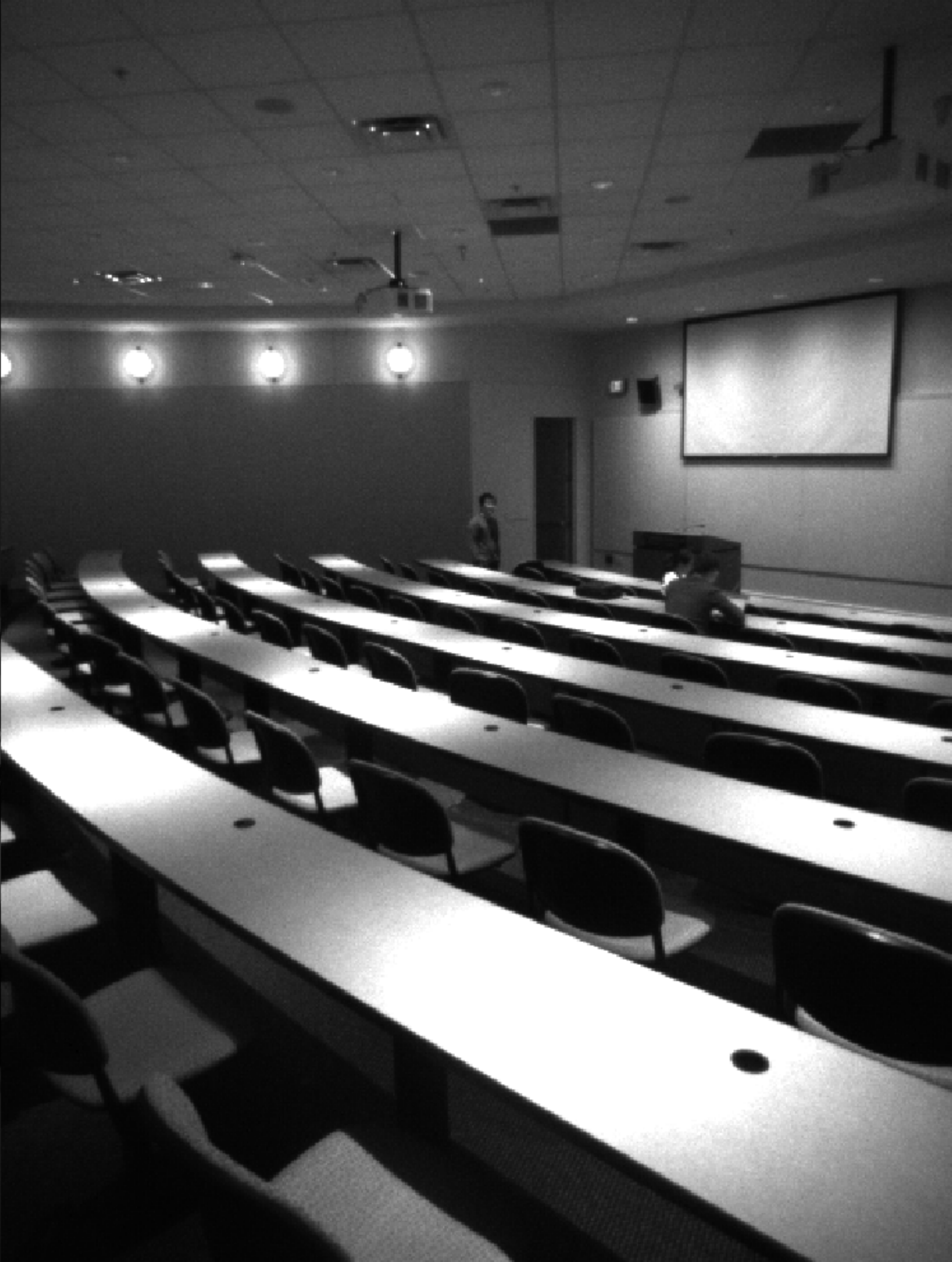}}
\\
\subfloat[KITTI \textit{Seq. 04}\label{fig:kitti04}]
        {\includegraphics[height=1in]{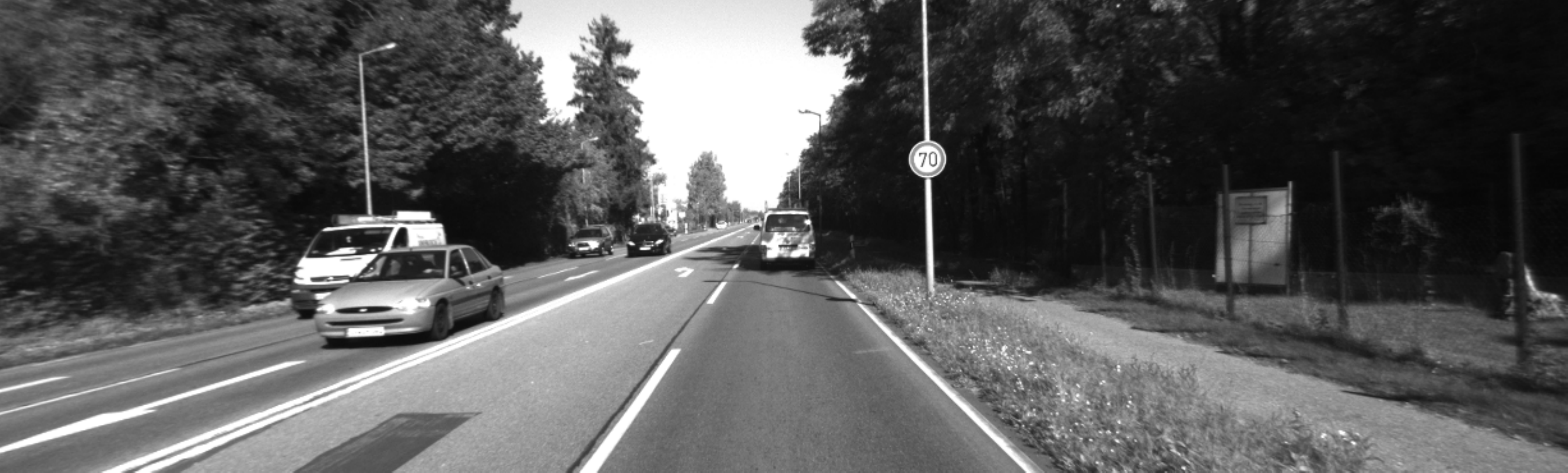}}
    \quad
\subfloat[EuRoC \textit{MH 05 difficult}\label{fig:mh05}]
        {\includegraphics[height=1in]{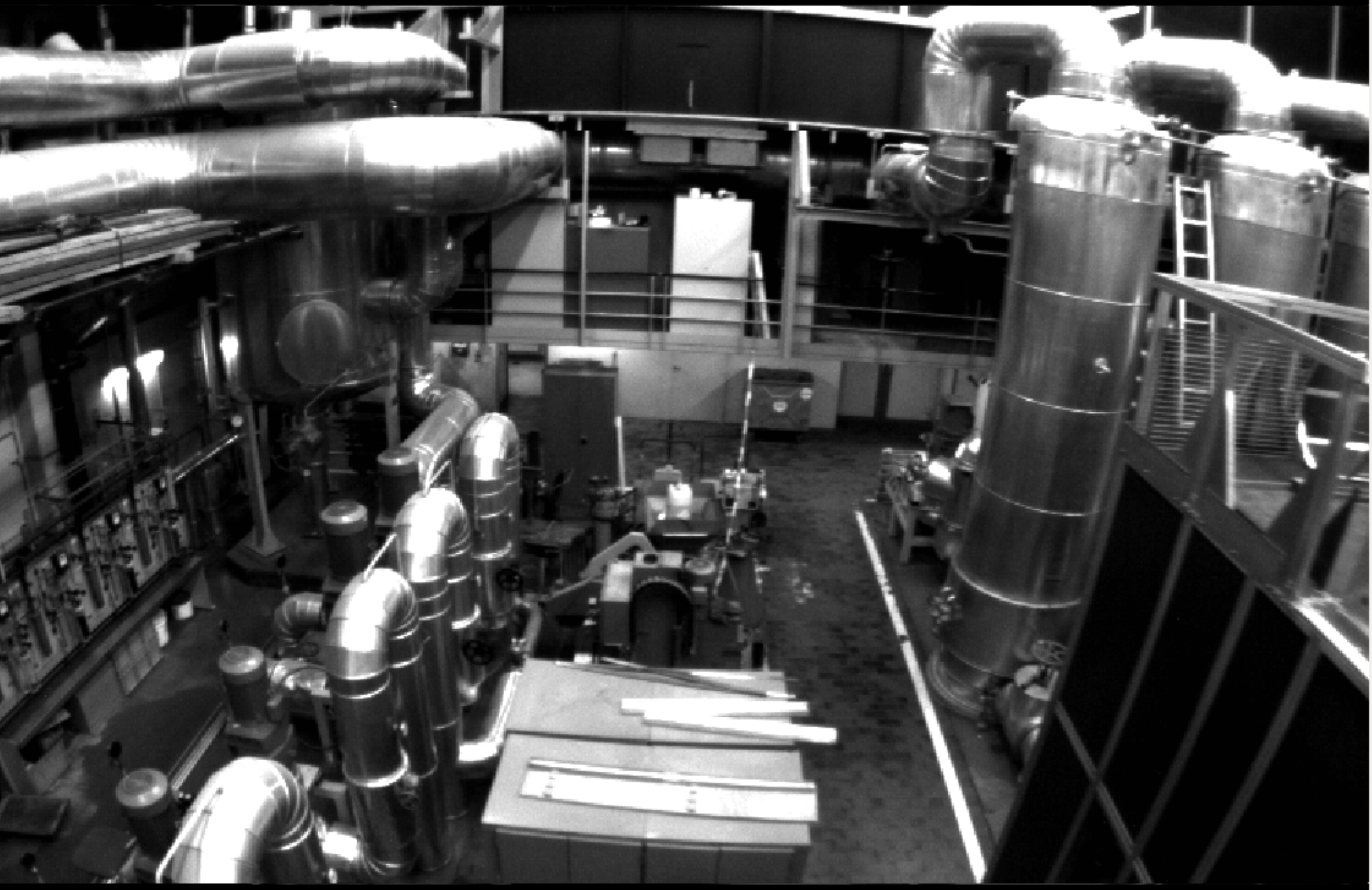}}
    \quad
\subfloat[EuRoC \textit{V1 03 difficult}\label{fig:v103}]
        {\includegraphics[height=1in]{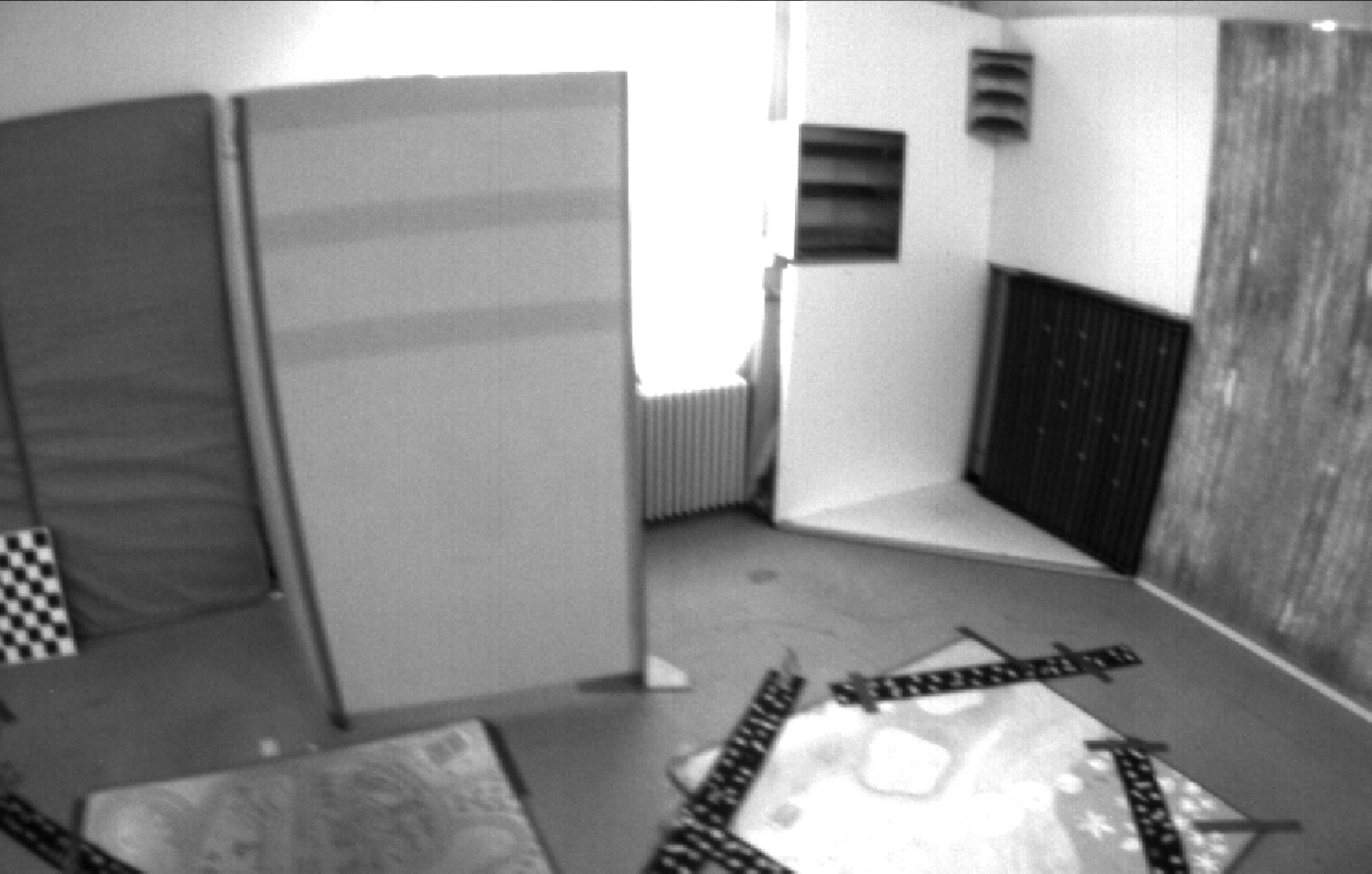}}
\\
\subfloat[NewCollege\label{fig:newcollege}]
        {\includegraphics[height=1in]{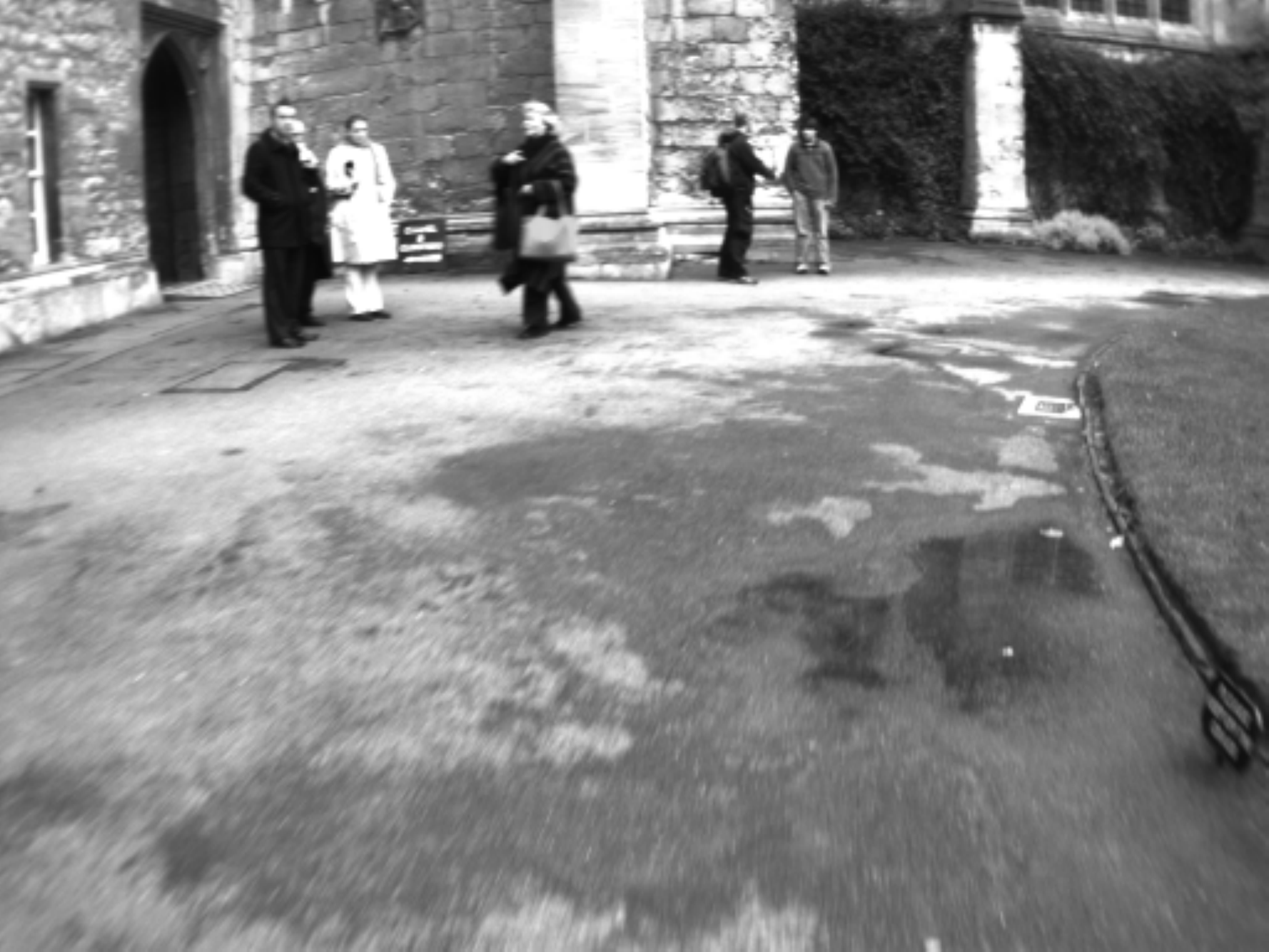}}
    \quad
\subfloat[TUM RGBD \textit{desk person}\label{fig:deskperson}]
        {\includegraphics[height=1in]{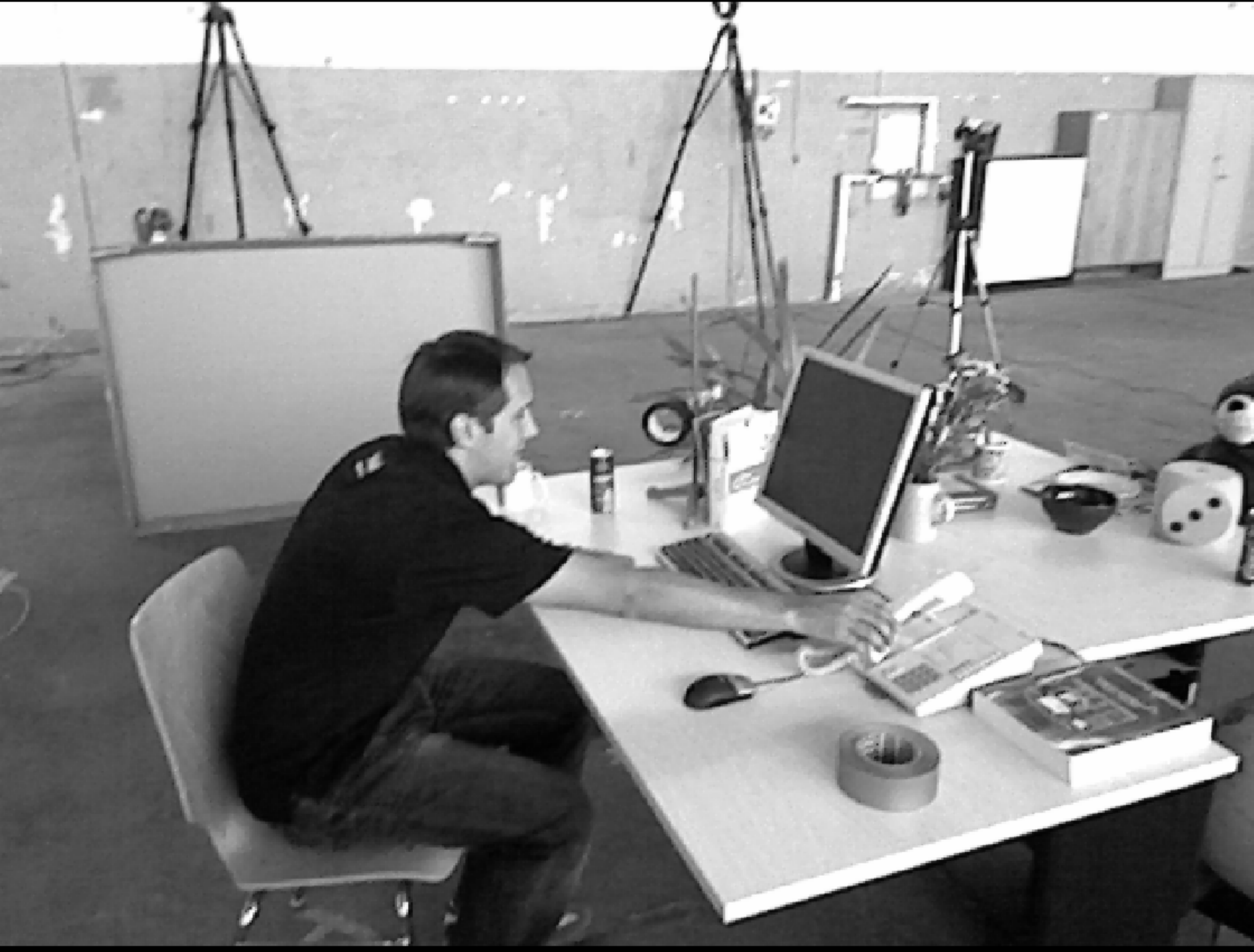}}
        \quad
\subfloat[\textit{Corridor} \label{fig:corridor}]
        {\includegraphics[height=1in]{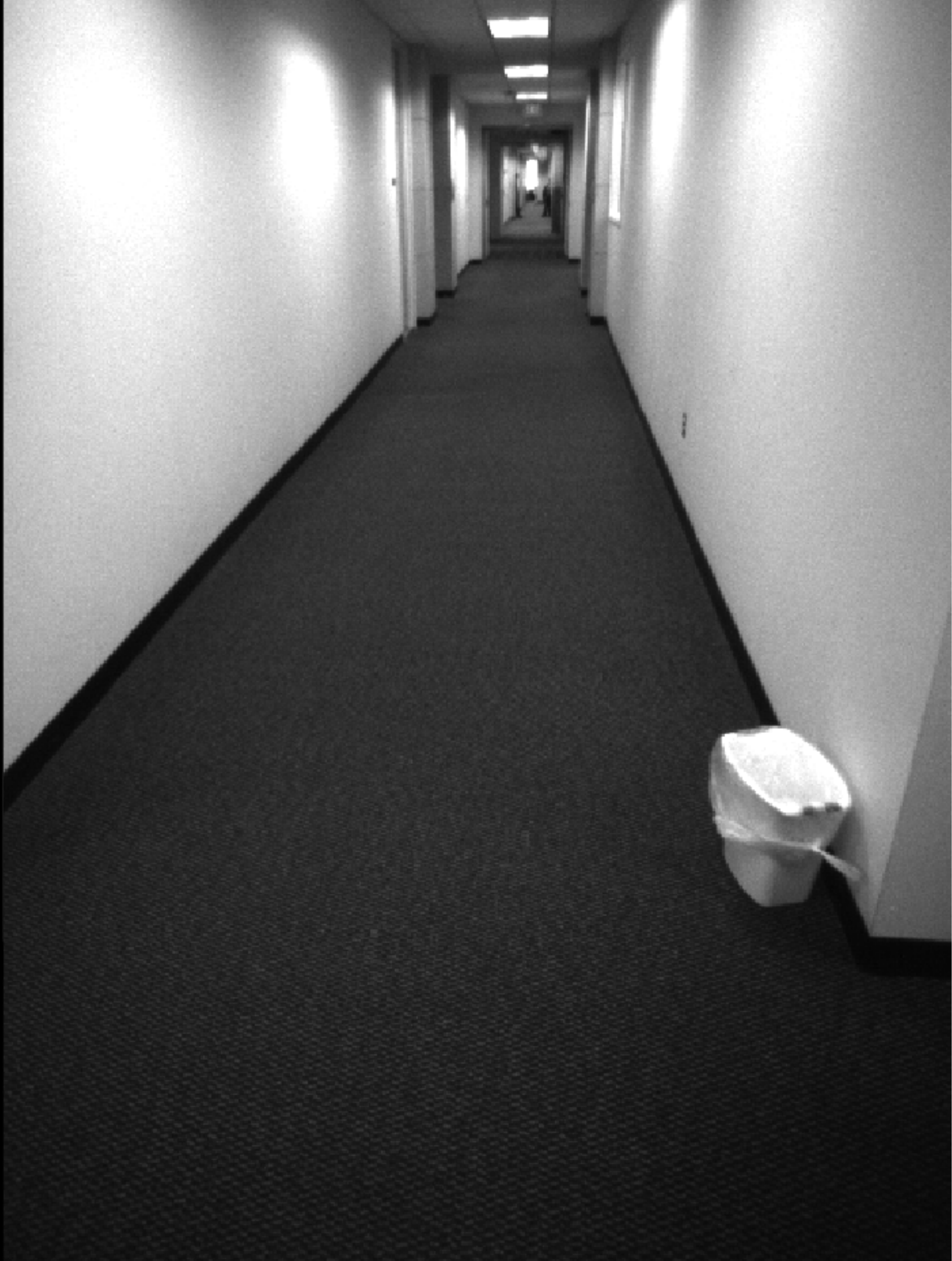}}
        \quad
\subfloat[ICL NUIM \textit{lr kt0} \label{fig:lr00}]
        {\includegraphics[height=1in]{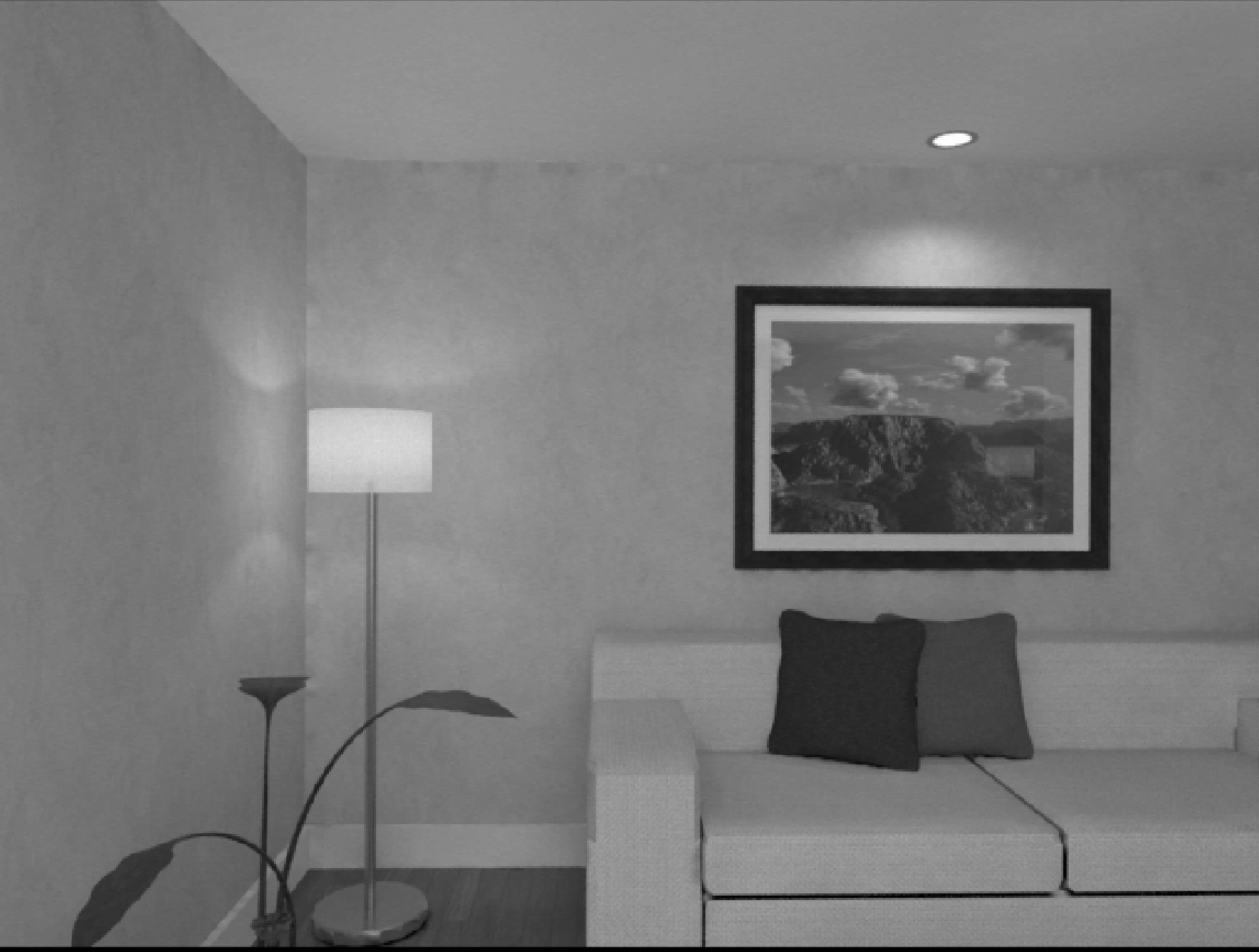}}
                \quad
\subfloat[ICL NUIM \textit{of kt3} \label{fig:of03}]
        {\includegraphics[height=1in]{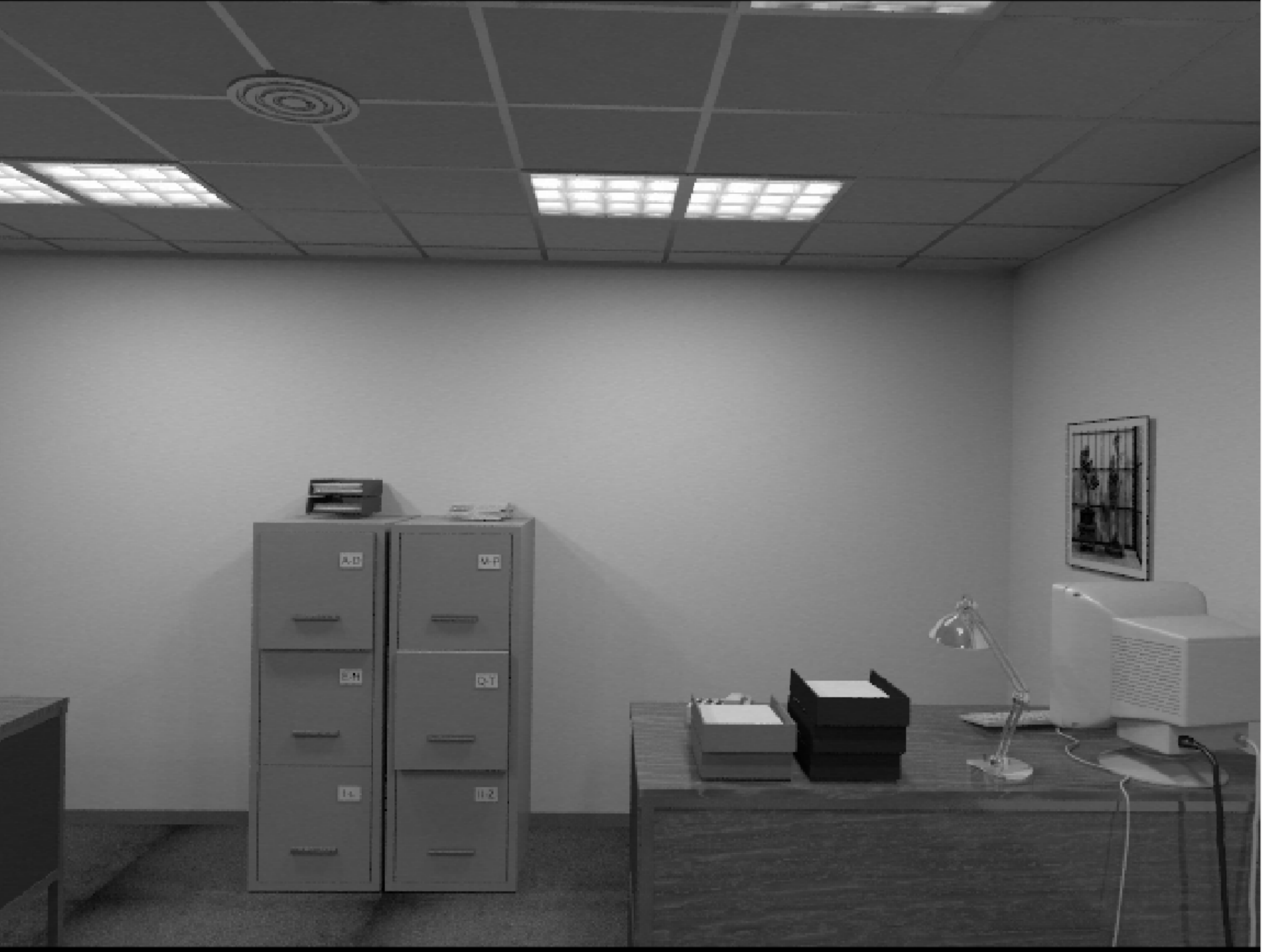}}
\caption{Characteristic images of selected typical sequences.} \label{fig:exampleimages}
\end{figure*}

For the analysis, we chose 
five
factors to serve as the
parameters of interest for characterizing and differentiating the
sequences.  They were scene, duration, environment dynamics, motion
dynamics and revisit frequency.  Some factors were merged into these
categories.  For example scene illumination and image exposure variations
were connected to the \textit{scene} attribute in order to have a more
manageable review workload.
Categorization for each properties is based on heuristic thresholds or
qualitative assessment. 
For example, the \textit{Duration} property is categorized as 
\textit{Short}, \textit{Long}, or \textit{Medium} if its duration
is below 2 mins, over 10 mins, or between the two thresholds, respectively. 
The \textit{Environ. Dynamics} is categorized as \textit{Low}, \textit{Medium} 
or \textit{High} if there are no or rarely moving objects in the scene, 
a few moving objects, or numerous or frequently seen object movements, 
respectively. 
Beyond those five properties, we also assigned difficulty labels to the
source benchmarks. When available in the source publication, we kept the
labeled assigned by the researchers. Otherwise, assignment was determined
using reported tracking outcomes or, if needed, through visual review of
the sequence. 
Table~\ref{tab:summ_vslam} provides sample descriptions of select
sequences, with individual frames from them shown in Fig~\ref{fig:exampleimages}. 
The complete version can be 
accessed online
.\!\!\footnote{\url{https://github.com/ivalab/Benchmarking_SLAM}} 
After reviewing the full table, we performed a downselection of the
benchmark sequences in order to balance the different categories.
Removal was determined by overlapping or common configurations or by
choosing a characteristic subset from a benchmark with many sequences.
The final set consisted of 
117
 sequences obtained from
various platforms (car, train, MAV, ground robot, handheld, head-mounted), 
scenarios (46\% indoor, 47\% outdoor, 7\% synthesized), 
duration (37\% short, 33\% medium and 30\% long), 
and motion patterns (33\% smooth, 38\% medium, 29\% aggressive).

\begin{figure}[th!]
  \centering
  \subfloat[Indoor + Outdoor\label{fig:dtree1_all}]
        {\includegraphics[width=0.80\columnwidth]{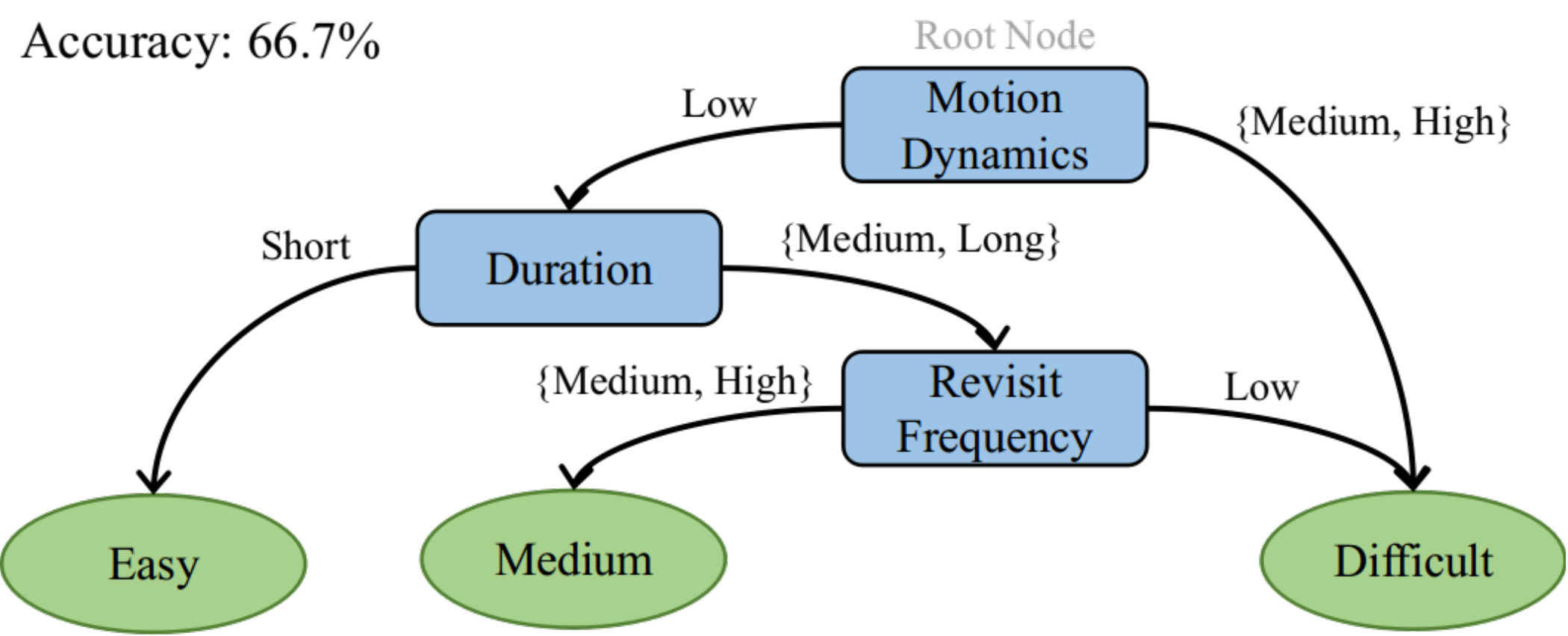}}
    \\
  \subfloat[Indoor Only\label{fig:dtree1_indoor}]
        {\includegraphics[width=0.80\columnwidth]{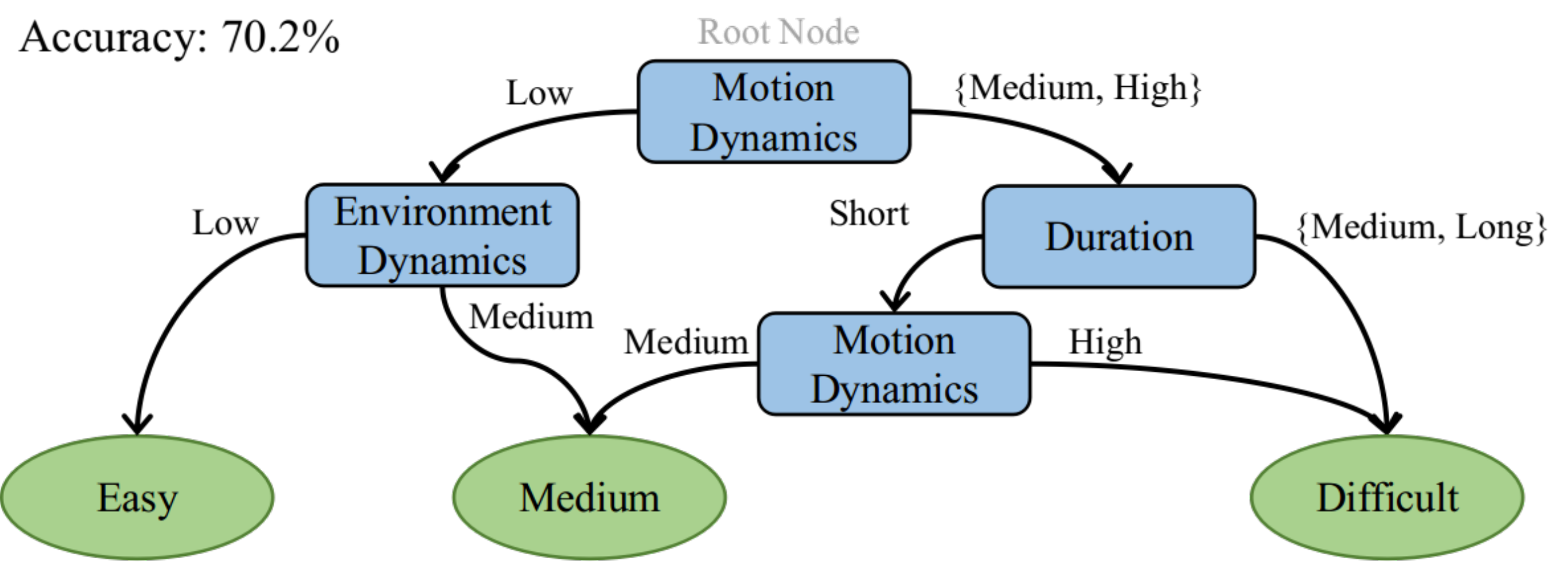}}
    \\
  \subfloat[Outdoor Only\label{fig:dtree1_outdoor}]
        {\includegraphics[width=0.80\columnwidth]{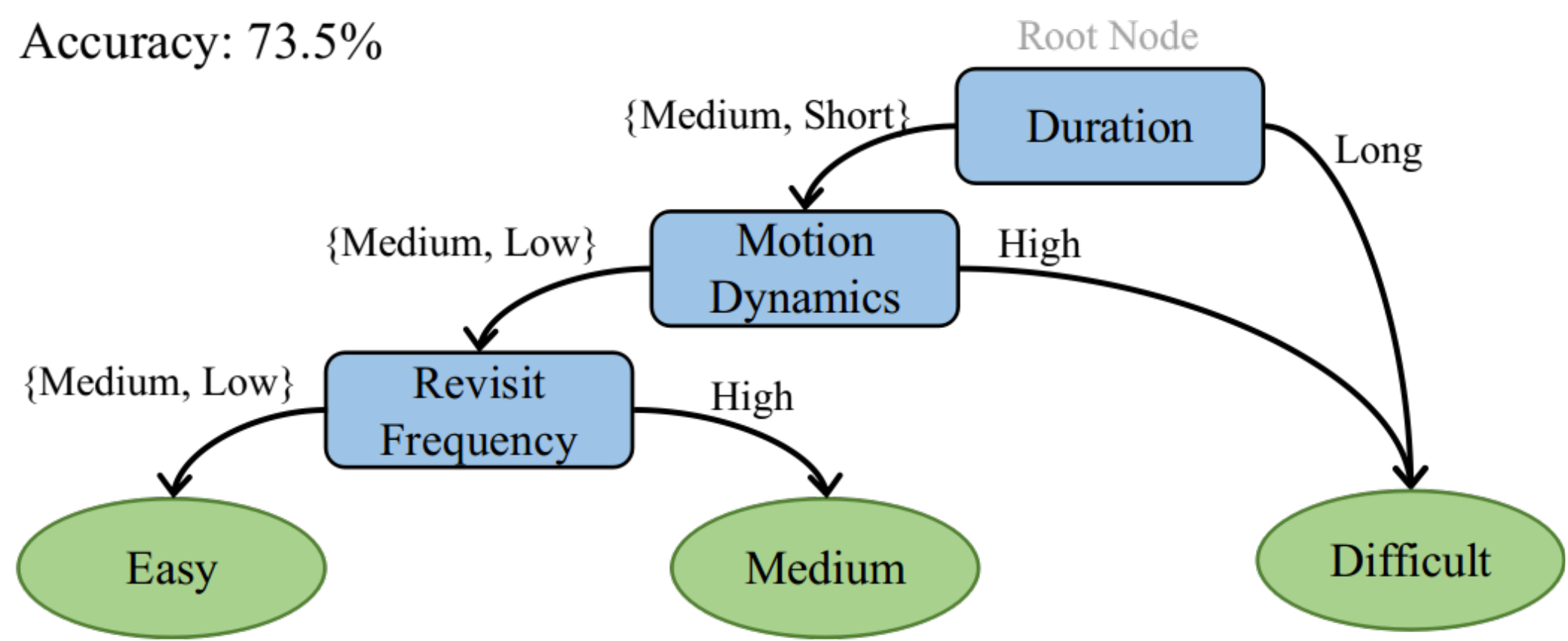}}
  \caption{Trained Decision Tree Factors influencing difficulty level.} \label{fig:dtree1}
  \vspace*{-2em}
\end{figure}

As a first pass at understanding what factors most impact the difficulty
annotation, we applied a decision tree classifier to the annotated set of
chosen benchmarks.  Each sequence is an observation with the 
five
 properties as the predictors and \textit{Difficulty}
as the response. 
Cross-validation is adopted in the process to examine the predictive
accuracy of the fitted models and meanwhile to protest against overfitting.
Specifically, the training pool is partitioned into five disjoint subsets,
and the training process is performed on each subset to fit a model, which
is trained using four other subsets and validated using its own subset.
The best model is adopted and re-assessed using the whole training 
pool to report an accuracy.
Performing the training procedure for the entire benchmark set, the
indoor-only subset and the outdoor-only subset leads to three decision
trees, all depicted in Fig~\ref{fig:dtree1} with prediction accuracy noted 
above and to the left of each tree. 

Common factors for all of the trees are \textit{Motion Dynamics} and
\textit{Duration}, with \textit{Motion Dynamics} being fairly consistent
regarding the final outcome. \textit{Duration} is also consistent across
the trees, however \textit{medium} durations evaluate differently between
indoor and outdoor datasets.
For indoor sequences \textit{Environment Dynamics} plays a role in
differentiating \textit{easy} versus \textit{medium}, whereas for outdoor
sequences it does not. It may reflect the different sensor hardware
associated to the two use cases (wide vs narrow field-of-view) and the
relative size of the moving objects within the image stream.
Interestingly the \textit{Revisit Frequency} has an opposing outcome for the
full dataset versus the outdoor dataset, suggesting the opposite role
of this factor for the indoor dataset though it is not a dominant one.
Revisiting for outdoor scenes may reflect the nature of loop closure at
intersections. There are four ways to cross an intersection but only one
crossing direction can trigger or contribute to loop-closure. For indoor
scenes with more freedom of movement, there may actually be less diversity
in view direction during revists.

Based on the decision trees, challenging sequences should be those with high
motion dynamics or long duration (irrespective of the motion dynamics).
To generate a reference set of sequences spanning these different decision
variables and reflecting distinct pathways, we reviewed the dominant
factors and identified 12 characteristic sequences across the three
performance categories.  
The \textit{easy} sequences are \textit{Seq 04}, 
\textit{lr kt0}, 
\textit{f2 desk person};
the \textit{medium} sequences are 
\textit{Conf. Hall1}, 
\textit{Seq 02}, 
\textit{room3}, \textit{of kt3}; and the
\textit{difficult} sequences are \textit{MH 05 diff}, \textit{V1 03 diff}, 
\textit{Corridor}, \textit{NewCollege}, \textit{outdoors4}.

\section{Time Profiling and Time Dilation}
\label{sec:timing}
\begin{table*}[t]
	\small
	\centering
	\caption{RMSE (M) / RPE (M/S) of 3 VO algorithms in slo-mo/normal speed on Selected Sequences}
	\begin{tabular}{c|c||c|c|c||c|c|c|c|c}
		\toprule[1pt]
		\midrule[0.1pt]
		\textbf{ } & 
		& \multicolumn{3}{c||}{\bfseries \small one run in \textit{slo-mo}}
		& \multicolumn{5}{c}{\bfseries \small five runs in normal speed (\#failures are highlighted by -/\textcolor{f4}{4}/\textcolor{f3}{3}/\textcolor{f2}{2}/\textcolor{f1}{1}/0)} \\
		\cmidrule{3-10}
		& \textbf{\small Seq.} & SVO\cite{SVO2017} & DSO\cite{DSO2017} & ORB\cite{ORBSLAM}  & SVO\cite{SVO2017} & DSO\cite{DSO2017} & ORB\cite{ORBSLAM} & GF-ORB\cite{zhao2018good2} & MH-ORB\cite{zhao2019maphash} \\
		\midrule
		\  \parbox[t]{1mm}{\multirow{8}{*}{\rotatebox[origin=c]{90}{\textbf{RMSE}}}} \ 
				
		& \textit{f2 desk person}  & 1.26e0 	& 1.25e-1	& \underline{4.76e-2}  & \textcolor{f1}{1.53e0}& \textcolor{f0}{5.36e-1}& \textcolor{f0}{5.94e-2} & \textcolor{f0}{3.51e-2} &  \underline{\textcolor{f0}{2.88e-2}}\\
		
		& \textit{lr kt0}			& 4.97e-1	& 2.21e-1	&	\underline{2.10e-1}	 & \textcolor{f0}{3.07e-1} & \underline{\textcolor{f0}{2.61e-1}} & - & - & -			\\		
		
		& \textit{of kt3}			& 6.43e-1	& \underline{3.82e-2}	&	5.58e-2		 &\textcolor{f0}{5.60e-1} & \underline{\textcolor{f0}{3.87e-2}}&\textcolor{f0}{2.57e-1} & \textcolor{f0}{2.76e-1} & 	 \textcolor{f0}{6.69e-2}	\\
		
		& \textit{room3}  			& 2.03e0	& 2.86-1	& \underline{2.09e-1}  & \textcolor{f2}{2.02e0}& - & \underline{\textcolor{f2}{1.80e-1}} & -&  -\\ 		
		
		& \textit{outdoors4}  		& -		 	& \underline{2.21e-2} 	& 8.39e-2  & \underline{\textcolor{f3}{1.74e0}}& - & - & -&  -\\ 
				
		& \textit{MH 05 diff}  		& 4.26e0 	& \underline{1.38e-1}	& 3.03e-1 &  \textcolor{f1}{1.44e0} & \underline{\textcolor{f4}{1.08e-1}}& \textcolor{f0}{1.18e0} & \textcolor{f0}{1.43e-1} & \textcolor{f0}{2.29e-1}\\ 
		
		& \textit{V1 01 diff}  		& 5.53e-1 	& 1.08e0 	& \underline{2.32e-1}  & \textcolor{f0}{6.09e-1} & \textcolor{f0}{1.34e0}& \textcolor{f3}{1.25e0}& \textcolor{f4}{9.23e-1}&  \underline{\textcolor{f2}{4.61e-1}}\\ 
		
		\cline{2-10} 
		\vspace{-1pt}
		& \textbf{Average}  		& 1.54 	& 0.50 	& \underline{0.16}  & 1.17 & 0.46 & 0.59 & 0.34 & \underline{0.20} \\ 
		\midrule
		\  \parbox[t]{1mm}{\multirow{5}{*}{\rotatebox[origin=c]{90}{\textbf{RPE}}}}\ 
				
		& \textit{KITTI Seq 04} 	& 2.06e0 	& \underline{7.88e-2} 	& 9.18e-2  & \textcolor{f0}{1.82e0} & \underline{\textcolor{f0}{8.09e-2}}&\textcolor{f0}{9.74e-2} & \textcolor{f0}{1.00e-1} &  \textcolor{f0}{9.92e-2} \\ 
		
		& \textit{KITTI Seq 02}		& 6.91e0		& \underline{1.31e-1}		& 1.38e-1	 & \textcolor{f0}{7.17e0} & \textcolor{f0}{1.52e-1}& \textcolor{f2}{2.08e-1} & - &  \underline{\textcolor{f3}{1.41e-1}}\\	
		
		& \textit{conf. hall1} 		& 4.35e-1  & \underline{4.33e-1}	& -	 & \textcolor{f0}{4.25e-1}& \textcolor{f0}{4.57e-1}& \underline{\textcolor{f4}{1.50e-1}}& \textcolor{f0}{1.64e-1}& \textcolor{f1}{2.17e-1}\\
		
		& \textit{corridor}			& 1.20e0	&	6.50e-1		&	\underline{2.34e-1}	 & \textcolor{f0}{1.38e0}& \underline{\textcolor{f0}{5.31e-1}}& \textcolor{f4}{1.54e0}& \textcolor{f0}{6.22e-1}& \textcolor{f1}{1.05e0}\\
		
		& \textit{NewCollege}  		& -        & 1.92e-2   & \underline{1.65e-2}  & - & \textcolor{f0}{1.93e-2} & \underline{\textcolor{f1}{1.88e-2}}& \textcolor{f0}{1.95e-2}&  \textcolor{f0}{1.92e-2}\\ 	
		\cline{2-10} 
		\vspace{-1pt}
		& \textbf{Average}  		& 2.65 	& 0.26 	& \underline{0.12}  & 2.70 & 0.25 & 0.40 & \underline{0.23} & 0.31 \\ 
		\midrule[0.1pt]
 	\bottomrule	[1pt]
	\end{tabular} 
	\label{tab:accuracy_summary}
\end{table*}

Time profiling of the computational modules of a SLAM system provides
clues to how SLAM implementations should be improved at the
computational component level.  This is particularly true for
feature-based methods, which typically are more costly than direct
methods.
To understand the time consumption of the modules in a SLAM 
pipeline, we advocate fine-grained time profiling and the use of
time-dilation when evaluating SLAM systems with ROS bag playback, i.e.
\textit{slo-mo} playback. 
The idea is similar to the \textit{process-every-frame} mode in 
SLAMBench~\cite{nardi2015introducing}, which continues with the next frame 
after the previous frame is completely processed, The proposed 
\textit{slo-mo} is straightforward and easy to apply in ROS. 
We conjecture that \textit{slo-mo} playback will establish performance 
upper-bounds for evaluated SLAM systems, which serves as a hint on the
potential of SLAM system (e.g. running on better hardware in the near
future).
The time scaling factor for \textit{slo-mo} playback was chosen to be
0.2, providing 5x more time for a single-frame update.

Quantitative eqvaluation on the chosen 12 sequences involved three 
state-of-the-art VO/SLAM algorithms, i.e. SVO~\cite{SVO2017},
DSO~\cite{DSO2017} and ORB-SLAM (ORB)~\cite{ORBSLAM}.  For those using
features, the feature quantity parameter was set to use 800 features 
per frame in \textit{slo-mo}. 
The testbed is a laptop with a Intel Core i7-6820HQ quadcore 2.70GHz 
CPU and 32 GB memory. 
The loop-closure thread in ORB-SLAM was disabled to operate like a
visual odometry (VO) system, though the local mapping thread was not
disabled (it behaves like a short-term loop closure). 
Each sequence was tested once for each SLAM algorithm.
Evaluation varied based on the available ground truth.
For sequences with high-precision 6DoF ground truth 
(e.g. from Motion Capture system), tracking accuracy is evaluated 
with RMSE (m) versus the absolute pose references.  
For sequences with less frequent ground truth signals or with
synthesized ground truth (e.g., using SfM), the RMSE of relative pose error 
(m/s) is used.  The time cost of the major computational components of
the three VO algorithms was recorded.

\begin{figure}[t]
	\centering
	\includegraphics[width=\columnwidth]{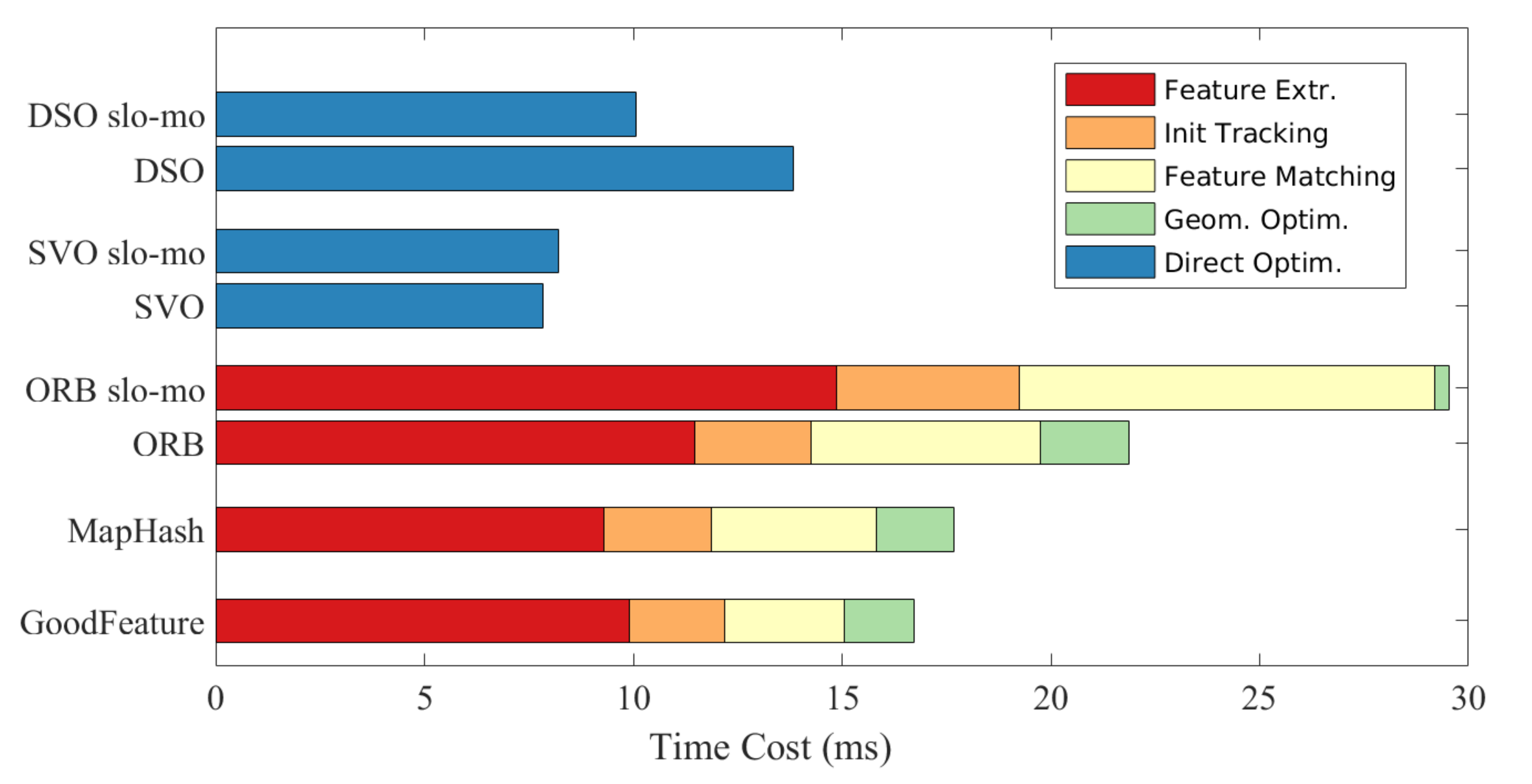} \\
	\caption{Time profiling of modules in three state-of-the-art VO algorithms and two low-latency algorithms, running under normal speed and \textit{slo-mo}.
 \label{fig:latency}}
\end{figure}

The timing outcomes for the tested algorithms are shown in
Fig~\ref{fig:latency}, where the estimated time cost for each component 
is computed by averaging over all tracked frames in all selected sequences. 
The methods with direct pose estimation components, SVO and DSO, did not
consume significantly more time. 
Interestingly, DSO ran faster in normal speed, which could due to 
the improved inverse-depth estimation provided by back-end.  
With these minor changes, weak performance points 
in these algorithms should be attributed to algorithm performance limits.
ORB-SLAM consumed more time in \textit{slo-mo} versus normal time for
many of the early components, but less time for the pose optimization
step. Faster convergence of the pose optimization implies better
conditioning of the optimization, better predicted poses, or improved
feature selection or coordinate estimation.
For total time cost, ORB-SLAM take the most time processing each frame,
primarily due to the feature extraction and matching.

Pose tracking performance results for \textit{slo-mo} are listed on 
the left side of Table~\ref{tab:accuracy_summary}. On each sequence, 
the method with the lowest RMSE/RPE is underlined and the failure cases 
with tracking loss over one third of the entire seuqence are discarded 
(marked as dash). 
Considering first track loss only, DSO is the only algorithm to
successfully track all sequences. Furthermore, it has good tracking
accuracy (second to ORB-SLAM). This strong performance suggests that
improvements to DSO will most likely involve additional components or
modifications outside of the core DSO components.
In terms of available tracking accuracy, ORB-SLAM achieves the best
performance with average RMSE of 0.16m and average RPE of 0.12m/s.  
However its timing does not match that of DSO, thus modifications should
prioritize enhancing ORB-SLAM's timing properties.
Though SVO has excellent timing, it has the lowest performance with
regards to track loss and pose tracking accuracy 
(average RMSE of 1.5m and average RPE of 2.65m/s).  
These outcomes indicate that, the \textit{slo-mo} can help understand 
performance properties of SLAM systems. 

\section{Dataset Properties Influencing Performance}
\label{sec:eval}
To explore the performance limits in actual operational conditions, the
\textit{slo-mo} results can be compared with the ones generated at
normal speed. Performance differences may point to potential source of
improvement by establishing modifications that nullify them.
We run these three algorithm at normal speed five times on each sequence.
We also applied two additional ORB-SLAM modifications that aim to lower
the compute time of the front-end computations 
\cite{zhao2018good2,zhao2019maphash} (time improvements can be seen in
Fig.~ \ref{fig:latency}).  The results are summarized on the right side of 
Table~\ref{tab:accuracy_summary}. 
To communicate tracking accuracy and the number of tracking failures, 
we compute the average tracking error only over successful cases, but 
mark the error in different gray levels according to failure quantity. 

Two of the three algorithms experience performance degradation to different
degrees when operating with time limit. One, SVO, did not significantly
change.  Though one additional sequence (\textit{outdoor4})was tracked
for one out of five runs, it did experience more failure than success.
Thus, we consider the change in track success rate to be negligible. The
tracking accuracy was within 2\% of the \textit{slo-mo} version. DSO
exhibits track loss for some sequences 
(\textit{room3}, \textit{outdoors4}, \textit{MH 05 diff}) relative to
\textit{slo-mo} which might also point to degradation of the back-end
processing due to the time constraints.  Further analysis would be
necessary to understand the source of these differences.
ORB degrades the most when returning back to normal speed, both in terms
of increased track failure and higher pose error. The higher time cost
of ORB-SLAM impacts performance, as ORB-SLAM has to skip frames to
complete the process initiated from an earlier one but not yet completed.
To examine the impact of lower latency two addtional low-latency algorithms, 
GoodFeature~\cite{zhao2019tro} and MapHash~\cite{zhao2019maphash} are
evaluated for comparison. 
Both were implemented in the ORB-SLAM framework and achieve low tracking
latency through different strategies. The former employs active matching
and the second employs more efficient local map subset selection. 
By lowering the pose tracking latency, the two algorithms improve
somewhat tracking success. A bigger improvement is seen for the pose
accuracy.

\begin{figure}[!t]
	\centering
	\includegraphics[width=0.9\linewidth]{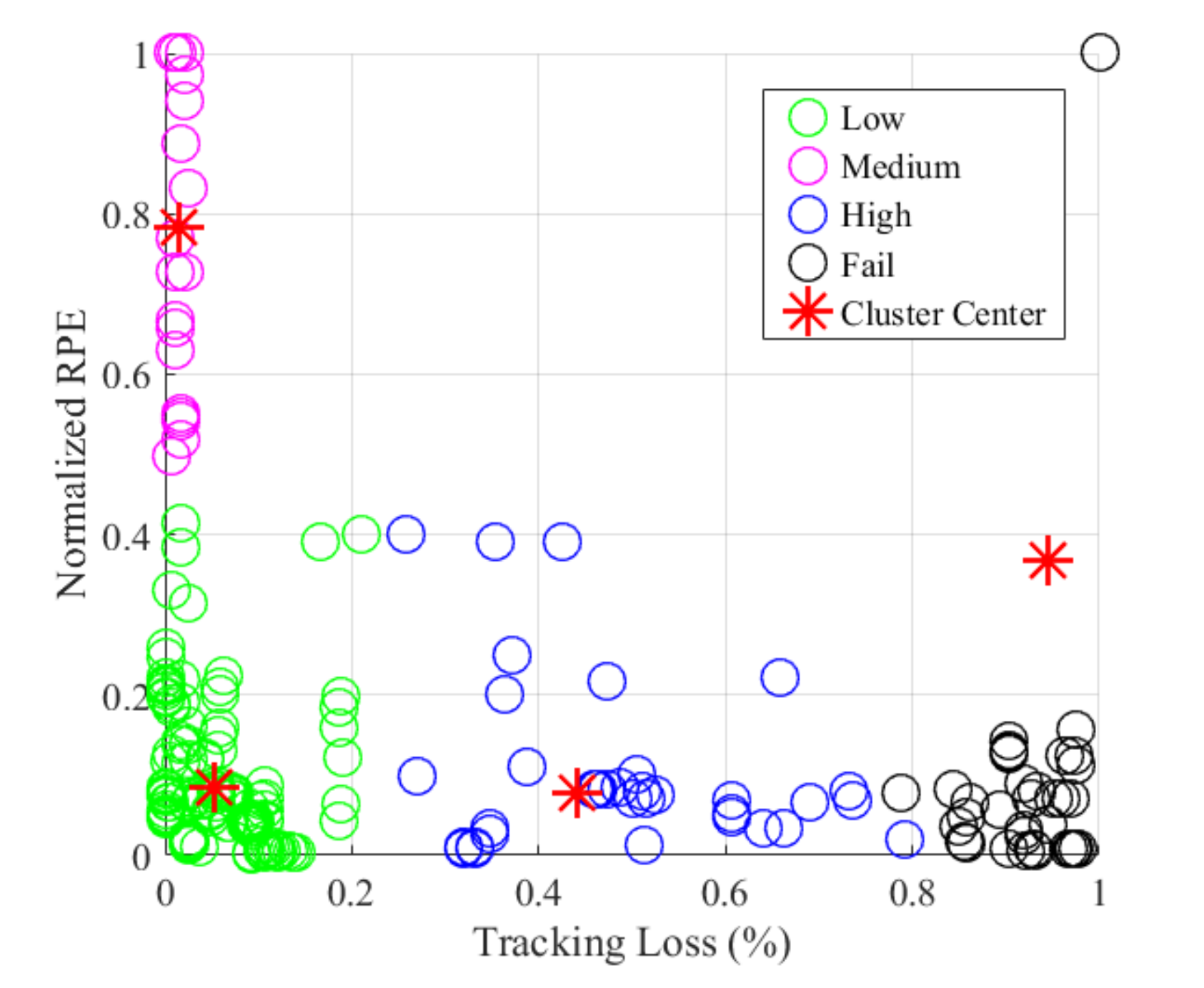} \\
	\caption{Kmeans++ clustering on pose tracking errors.
 \label{fig:clustering}}
\end{figure}

\begin{figure*}[th!]
  \centering
\subfloat[DSO\label{fig:retrain_dso}]
        {\includegraphics[width=0.8\columnwidth]{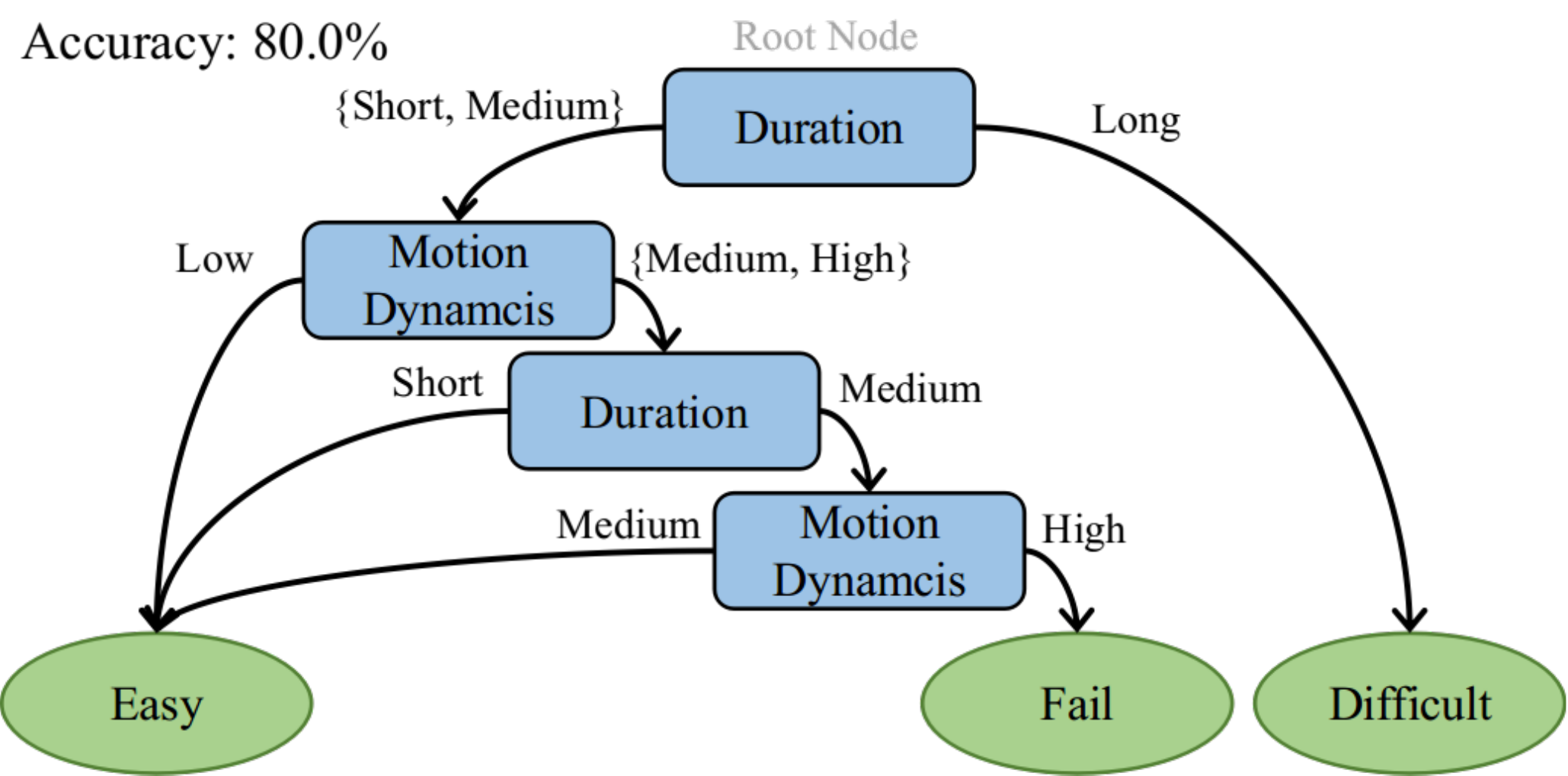}}
    \qquad
\subfloat[SVO \label{fig:retrain_svo}]
        {\includegraphics[width=0.8\columnwidth]{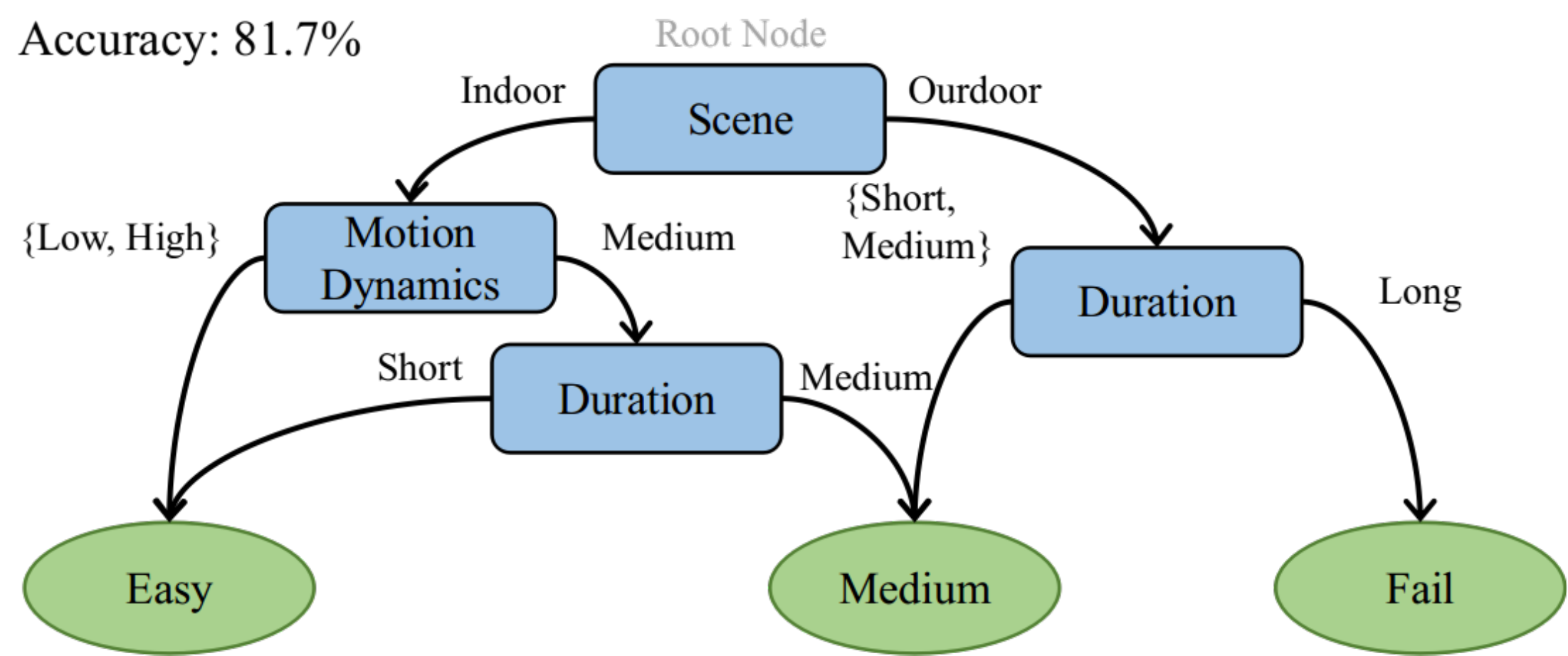}}
\\
\subfloat[ORB \label{fig:retrain_orb}]
        {\includegraphics[width=0.8\columnwidth]{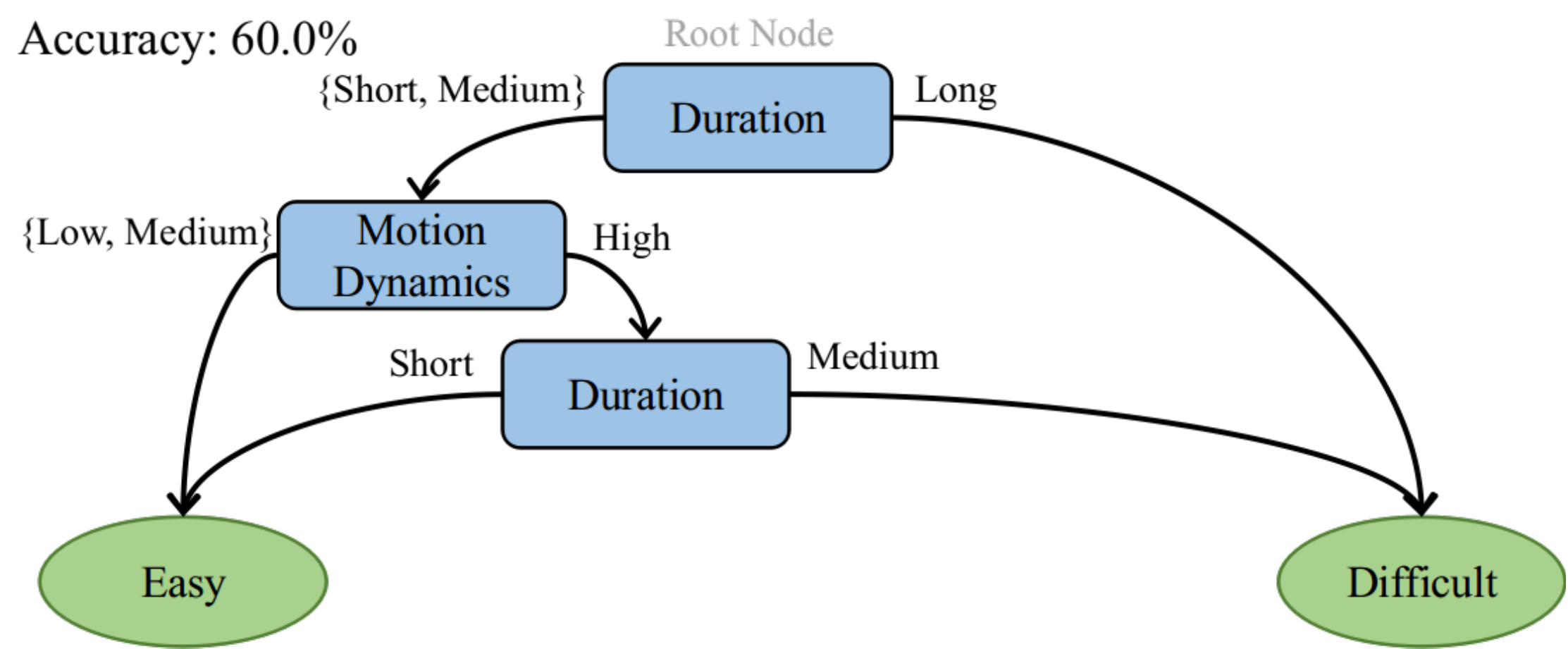}}
            \qquad
\subfloat[ALL \label{fig:retrain_all}]
        {\includegraphics[width=0.8\columnwidth]{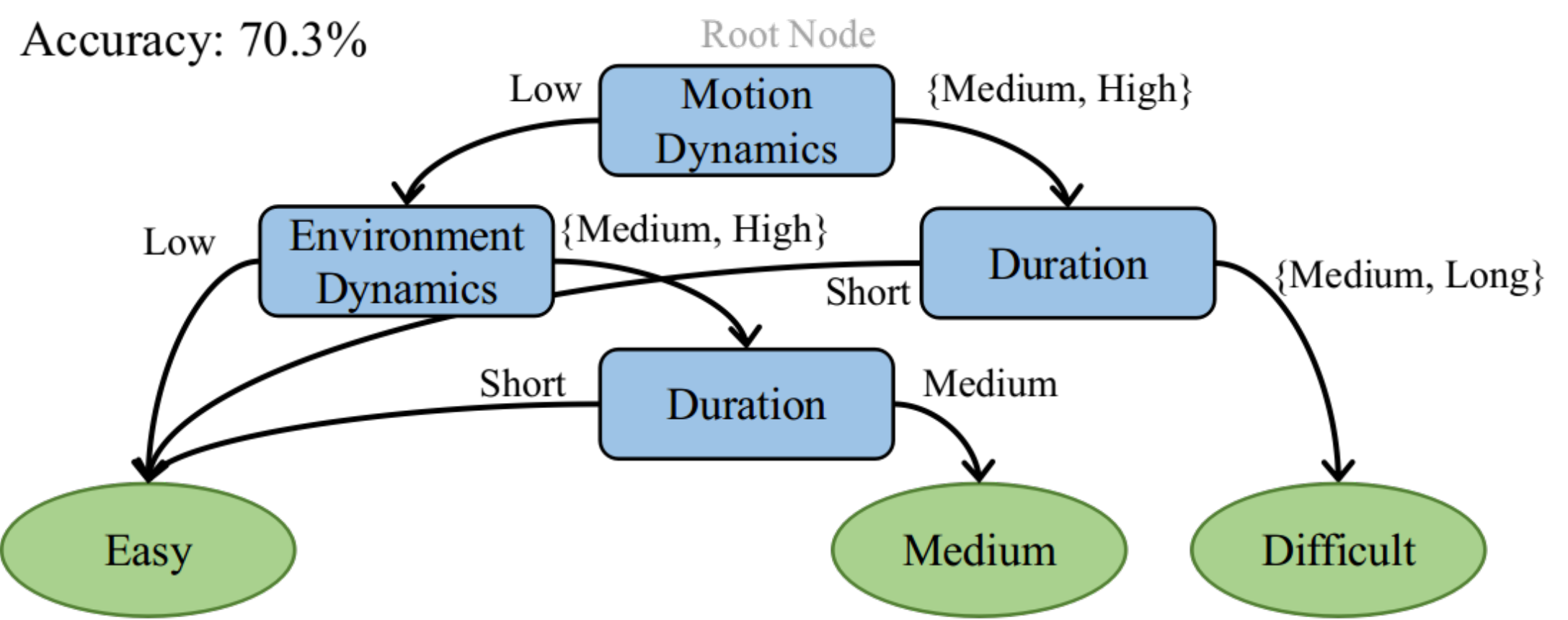}}
\caption{Trained decision trees using pose tracking errors.} \label{fig:finaltrees}
\end{figure*}

While it is important to understand how well given visual SLAM algorithm
work in terms of relative standings, a better understanding or
characterization of performance would illuminate where additional effort
should be spent improving a particular SLAM algorithm. 
Here, we replicate the exploration of benchmark properties of 
Section~\ref{sec:bench} but use the quantitative outcomes from the
selected sequences.  In particular, we re-annotate the
\textit{Difficulty} label based on the track loss rate and the tracking
accuracy.
Each run with each algorithm on each sequence is taken as an observation, thus 
there will be 300 two-dimensional observations in total for training. 
To prevent biasing, we saturate RPE at 2 m/s and normalize the values. 
These two factors yield four candidate categories. 
An algorithm performance is considered as \textit{high} if it can track poses 
with low loss and low RPE. 
If it tracks the entire sequence but with poor accuracy, we consider
performance to be \textit{medium}. 
Moreover, we mark the performance as \textit{difficult} if it fails to track 
sometime in the middle of the sequence, and mark them as \textit{fail} if it 
is lost in the begining, no matter how accurately it tracks. 
K-means++\cite{arthur2007k} is applied to cluster observations into these four categories. The distribution of observations and the clustered centroids are 
shown in Fig~\ref{fig:clustering}. 

Given the cluster results, we categorize the pose tracking performance and 
build a decision tree for the three main SLAM algorithms tested. 
Since the tracking results are generated only from three VO algorithms,
the property \textit{Revisit Frequency} is removed as a factor.
The tree strucuture, from top to bottom, can indicate the significance
of sequence properties to each algorithm. 
According to Fig~\ref{fig:finaltrees}, these algorithms act differently
in terms of the selected characteristic sequences. 
For SVO, \textit{Scene} is the most important factor for decent operation, 
and \textit{Motion Dynamics} and \textit{Duration} come in the second place. 
No \textit{Difficult} leaf node exists, meaning that SVO usually tracked all 
the way if it is successfully started. 
The \textit{Medium} leaf node is connected with two \textit{Duration}
middle nodes, indicating its tracking accuracy depends to a great extent
on the sequence length. 
An interesting insight for DSO and ORB is their similarity. 
Their upper structures are basically the same, from \textit{Duration}, \textit{Motion Dynamics} to \textit{Duration}, which might partially explain the reason they can obtain competitive tracking accuracy on selected sequences.
The difference lies in the last judgement for \textit{Motion Dynamics}, where 
DSO can handle \textit{Medium} dynamics but will fail when it is \textit{High}. 
In contrast, ORB is not affected by \textit{Motion Dynamics} at this point, 
and no related failure will be caused. 
In summary, all algorithms as sensitive to \textit{Duration}, which is 
related to map maintenance, environment changes, and drift correction if
a loop closure module is available. 

In addition to the algorithm-specific decision trees, we aggregated all
of the data to generate a decision tree. 
Similar pre-process, clustering and training steps were conducted with the 
entire set, but with clustering into three categories. 
The generated decision tree is displayed on the lower right of
Fig~\ref{fig:finaltrees}. 
This tree is a quantitative vesion of the tree from Section~\ref{sec:bench}, 
based on actual outcomes as opposed to subjectively determined labels.
By comparison we can find that, both trees take the \textit{Motion Dynamics} as 
the first important factor, then comes \textit{Duration} and other factors. 
Our knowledge about what SLAM can do and what scenarios it can complete
is consistent with the quantative truth. 
A more comprehensive understanding can be obtained if breaking the
sequence properties into finer scale, but this comparion presents at
least two promising fields that SLAM research can focus on in the near
future: 
1) the robustness under aggressive motion patterns and 
2) the ability to handle long-term operation.
The former is usually handled by visual-inertial SLAM methods, to which
the same analysis can be applied.  The latter will require developing a
quantitative analysis methodology to better establish how to improve
long-term operation.

\section{Conclusion}
This paper characterizes state-of-the-art SLAM benchmarks 
and methods, with special attention on challenging 
benchmark properties and crucial components 
within the SLAM pipeline.  A decision tree is proposed 
to identify these properties and components.  By comparing the performance efficiency 
of SLAM systems on both normal speed and slo-mo playback, we are able 
to identify how SLAM implementations should be improved at the 
computational component level, and suggest where future research effort
should be dedicated to maximize impact.

\balance 
\bibliographystyle{IEEEtran}
\bibliography{../../full_references}

\end{document}